\documentclass[letterpaper, 10 pt, conference]{ieeeconf}  

\IEEEoverridecommandlockouts                              

\title{\LARGE \bf
LiStereo: Generate Dense Depth Maps from LIDAR and Stereo Imagery
}

\author{Junming Zhang$^{1}$, Manikandasriram Srinivasan Ramanagopal$^{2}$, Ram Vasudevan$^{3}$ and Matthew Johnson-Roberson$^{4}$%
\thanks{
$^{1}$J. Zhang is with the Department of Electrical Engineering and Computer Science, University of Michigan, Ann Arbor, MI 48109 USA {\tt\small junming@umich.edu}
\newline \indent $^{2}$M. Srinivasan Ramanagopal is with the Robotics Program, University of Michigan, Ann Arbor, MI 48109 USA {\tt\small srmani@umich.edu}
\newline \indent $^{3}$R. Vasudevan is with the Department of Mechanical Engineering, University of Michigan, Ann Arbor, MI 48109 USA {\tt\small ramv@umich.edu}
\newline \indent $^{4}$M. Johnson-Roberson is with the Department of Naval Architecture and Marine Engineering, University of Michigan, Ann Arbor, MI 48109 USA {\tt\small mattjr@umich.edu}}
}

\usepackage{perl_acronyms}
\usepackage{graphicx}
\usepackage{amsmath}
\usepackage{tabularx}
\usepackage{float}
\usepackage{xcolor}
\usepackage{graphicx}
\usepackage{flushend}
\usepackage{textcase}
\usepackage[tablename=TABLE]{caption}
\DeclareCaptionTextFormat{up}{\MakeTextUppercase{#1}}
\usepackage[caption=false]{subfig}
\usepackage[
    style=ieee,
    doi=false,
    isbn=false,
    url=false,
    eprint=false,
    backend=bibtex,
    natbib=true
    ]{biblatex}

\newcommand{\specialcell}[2][c]{%
  \begin{tabular}[#1]{@{}c@{}}#2\end{tabular}}
\usepackage{multirow}
\usepackage{color}

\newcommand{\matt}[1]{}
\renewcommand{\matt}[1]{{\color{red} MattJR: {#1}}}
\newcommand{\katie}[1]{}
\renewcommand{\katie}[1]{{\color{red} Katie: {#1}}}
\begin{document}

\maketitle

\begin{abstract}
    An accurate depth map of the environment is critical to the safe operation of autonomous robots and vehicles. Currently, either light detection and ranging (LIDAR) or stereo matching algorithms are used to acquire such depth information. However, a high-resolution LIDAR is expensive and produces sparse depth map at large range; stereo matching algorithms are able to generate denser depth maps but are typically less accurate than LIDAR at long range. This paper combines these approaches together to generate high-quality dense depth maps. Unlike previous approaches that are trained using ground-truth labels, the proposed model adopts a self-supervised training process. Experiments show that the proposed method is able to generate high-quality dense depth maps and performs robustly even with low-resolution inputs. This shows the potential to reduce the cost by using LIDARs with lower resolution in concert with stereo systems while maintaining high resolution.
\end{abstract}

\section{INTRODUCTION}
    Depth estimation is one of the fundamental tasks in the computer vision field. Being able to acquire accurate depth maps is a first step for many vision tasks, such as 3D object detection~\cite{ku2018joint}, 3D mapping~\cite{zhang2015visual} and localization~\cite{wolcott2014visual}. These tasks underlie a variety of applications including augmented reality, autonomous driving, and robotics. 

    One way to construct these depth maps is by solving the stereo matching problem. 
    Traditionally, this has been done by applying window-based methods~\cite{adhyapak2007stereo} or global-optimization methods~\cite{semiglobal} to construct a disparity map. Recently, deep convolution neural networks (DCNNs) have been applied to solve the stereo matching problem~\cite{zbontar2016stereo, mayer2016large}, since they are empirically verified to perform better than traditional approaches on a variety of vision tasks such as image classification~\cite{resnet} and object detection~\cite{MaskRCNN}. Generally, DCNNs are trained end-to-end with a large amount of ground-truth labels. Using sophisticated network architectures and ground-truth disparity labels, DCNNs have achieved impressive results as shown on KITTI benchmark stereo matching task~\cite{menze2015object}. However, there are still some important challenges to be addressed. It is difficult to find correspondence in regions of high specularity, low-texture or under  occlusion. In addition, the reliability of the estimated disparity map is range-dependent, which means that disparity estimated on distant regions tend to be less reliable than regions closer to the camera.

    LIDAR, in contrast, is able to accurately measure depth over long ranges. The most common form of LIDAR relies on emitting pulses of light and measuring the time it takes those pulses to reflect off objects in the environment and return to the sensor. The primary limitation of such sensors is their angular resolution. This resolution is determined by the number of beams and is a function of the number of receivers. Since the cost to increase the number of beams is prohibitive, LIDAR typically generates sparse point clouds. This is especially problematic for distant objects. The depth map generated by stereo correspondence is much denser than one obtained from a LIDAR. But small errors in matching lead to large errors in range for distant objects.
    Previous research has exploited the high angular resolution of RGB images and use it to upsample the sparse depth map~\cite{ferstl2013image, maSelf,van2019sparse,spatialProp}. The aim of these approaches is to calculate depth values at the high angular sampling rate of an RGB image by interpolating the empty pixels in the sparse depth map. Typically, ground-truth depth maps are required during training. With this pipeline, recent papers achieve impressive results on the KITTI benchmark depth completion task~\cite{sparsityCNN}. Still, extrapolating in large regions without LIDAR depth remains an open problem. Additionally, collecting a large amount of ground-truth depth labels for training is expensive.
    
  This paper generates dense depth maps by utilizing stereo images and sparse depth maps collected from a LIDAR. We demonstrate that the stereo cues are more powerful than monocular color guidance, as they provide direct estimates of depth. The contributions of this paper are:
    
    1. We propose a model that takes stereo images and LIDAR derived sparse depth maps as inputs and outputs accurate dense depth maps.

    2. The proposed model can be trained in a self-supervised manner, which avoids the cost of collecting a large amount of ground-truth labels. 

    3. Experiments show the advantages of stereo cues over monocular images in terms of sparser input depth maps and the potential to reduce the cost by using LIDARs with lower resolution.

\section{RELATED WORK}

    \textbf{Stereo Matching}: In recent years, typical stereo matching pipelines have been gradually replaced by end-to-end training DCNNs. In DCNNs, stereo features extracted from a deep Siamese structure are passed into a correlation module. Disparity maps are typically derived from several layers of convolutional matching. For supervised approaches, the models are trained using ground-truth disparity maps. There are many works focusing on efficiently forming a correlation volume~\cite{zbontar2016stereo,luo2016efficient,kendall}, designing architectures to extract features~\cite{PSMnet,pang2017cascade,flownet2,spatialProp} and resorting to extra information to refine results~\cite{segstereo,dissegnet}. Currently, the top methods on the KITTI benchmark stereo matching task are able to achieve an error rate around 1.5\%. In addition to supervised approaches, unsupervised methods have gained popularity~\cite{dissegnet,unsupervisedGeometry,godard2017unsupervised}. Unsupervised approaches rely on a warping error to provide a training loss. The warping error measures the difference between the warped image from one-side of the stereo pair and the input image from the other side. The warping process is implemented using differential bilinear interpolation~\cite{spatial}. Still, there is a noticeable gap between the supervised methods and unsupervised methods with respect to performance. Incorporating warping loss in supervised training has been shown to be beneficial over supervised training alone~\cite{semi-supervisedStereo}.

    \textbf{Depth estimation}: Depth estimation is one of the fundamental tasks in computer vision. There is an extensive body of research that has been done on depth estimation from a single color image and the error rate of this task has been reduced significantly in recent years~\cite{saxena2006learning,saxena20083,eigen2014depth,DeepOrdinal,li2018deep}. Models in those approaches learn to exploit the 2D monocular cues to predict depth. However, predicting depth from a monocular image is not a fully-constrained problem. The current results are not accurate enough for practical deployment in applications such as motion planning. The 2D monocular cues, such as occlusion and perspective, are more useful in relative depth prediction~\cite{xian2018monocular, chen2016single}. 
    
    The depth completion task is a sub-problem of depth estimation. Instead of knowing nothing about the scene, the depth completion task has strong priors on scene depth.  Sparsity-invariant operations have been developed for this task and have proved to be more effective than regular convolutions~\cite{sparsityCNN,HmsNet}. With additional color images, the depth completion process can be guided by color information. Recent works \cite{chen2018estimating,Sparse2Dense,dfusenet} show a performance boost using the color information contained in RGB data. Regular convolutions turn out to be successful in learning depth completion if dense color images are provided~\cite{jaritz2018sparse}. 
    Additionally, there are other approaches that focus on exploring extra information, such as edge and semantic cues~\cite{semanticallyDepth} and temporal frames~\cite{maSelf}. During training, camera poses are estimated between frames and the transformation matrix is used to warp images from one frame to the next~\cite{maSelf}. This warping loss enables self-supervised training. In \cite{park2018high}, precomputed disparity maps and LIDAR derived depth maps are taken as inputs and end up with high-resolution disparity maps.
    
    Specifically, our proposed method is intended for the depth completion task. Both \cite{maSelf} and our proposed method use warping loss to enable the model to be trained in a self-supervised manner. Unlike \cite{maSelf}, we explore stereo geometry instead of temporal information. With time-synchronized stereo images, observed position can be attributed to parallax while temporal cues suffer from additional ambiguity as observed position can be attributable to both parallax and object motion. 
    In essence, temporal frames just improve model's ability to exploit monocular cues for training, as they are not used during inference. 
    This paper proposes a model which takes stereo images and sparse depth maps as inputs and outputs dense depth maps. In addition this proposed model, which we call LiStereo (a portmanteau for LIDAR and Stereo Images), can be trained in a self-supervised manner.

\begin{figure*}[t!]
    \centering
    \includegraphics[width=0.8\linewidth]{{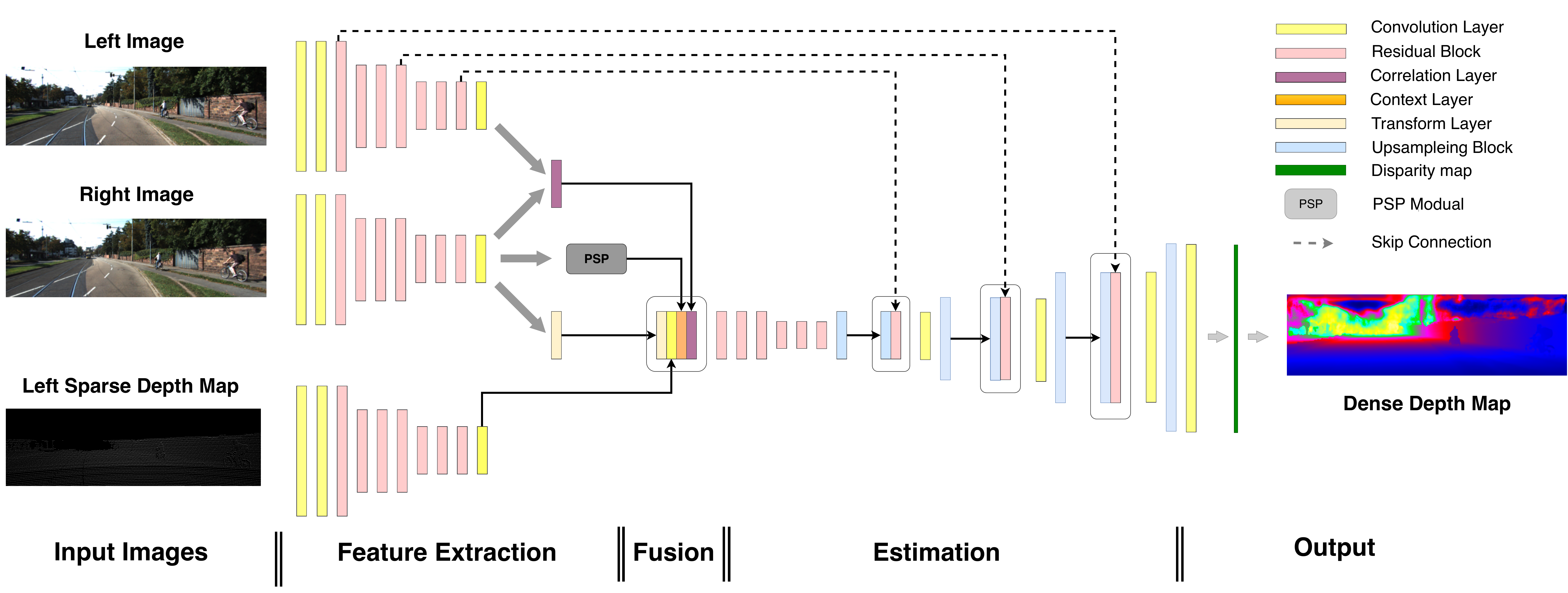}}
    \caption{Architecture of our proposed model. The pipeline of our model consists of following parts. (a) Inputs: rectified stereo images and corresponding left sparse depth map. (b) Feature extraction: features are extracted from stereo images and sparse depth map. The correlation layer computes correlation from one side of view to the other. Features from left color image are processed by transform layer to prepare for later sensor fusion. The PSP module is used to incorporate more contextual information. (c) Fusion: correlation information and features from the depth map are fused by concatenation. (d) Estimation: fused information is processed to perform depth estimation. (e) Output: a dense depth map and disparity map are generated. The output depth map is colorized for better visualization.}
    \label{architecture}
\end{figure*}
\vspace{-3mm}

\section{METHOD}
We present a model which generates dense depth maps by utilizing stereo images and sparse depth maps  collected from a LIDAR. The whole architecture of our model is illustrated in Figure.~\ref{architecture}.

\subsection{Architecture}
    There are two branches, the color image branch and the LIDAR branch. To extract features from stereo images, we borrow the Siamese structure in \cite{dissegnet}. The Siamese structure contains a ResNet50 structure \cite{resnet} and generates high-level features for both left and right images. Those feature maps are 1/8 of original input image size and are then processed to form a cost volume through the use of a correlation layer, similar to~\cite{mayer2016large}. The correlation layer correlates features from the left and right view horizontally. In the experiments, we consider a maximum displacement of 24 pixels in the feature map, which corresponds to maximum disparity of 192 pixels in the input image. The PSP module~\cite{zhao2017pyramid} is used to extract more contextual information. The branch for processing sparse depth maps has the same structure as the Siamese network, but it does not share parameters. Instead of summing them together, we fuse the information by concatenating the cost volume, the context layer, the transformed layer and LIDAR features, and pass them to six residual blocks. The output from this is followed by four upsampling blocks in order to output depth maps at the same resolution as the input images. Injecting skip layers from the left image branch of the Siamese structure into later layers is helpful to maintain high resolution information from the color image. 
    The output layer has 193 channels corresponding to the maximum disparity of 192 pixels, with a channel for zero disparity. Soft argmax~\cite{kendall} is used to generate the final disparity maps, and the absolute depth is then computed given the extrinsic parameters of the stereo cameras.

    In detail, the convolution layer in this paper refers to a convolution layer followed by a batch normalization layer~\cite{batchnorm} and a leaky ReLU rectifier~\cite{xu2015empirical}. The output layers of the Siamese structure, LIDAR branch and final output layer contain a regular convolution only. The size of all kernels is 3 except for the first two convolution layers in the Siamese structure and LIDAR branch, which are 7 and 5 respectively. The residual block is three-layer deep and the upsampling block consists of a bilinear interpolation layer, a regular convolution layer, a batch normalization layer and leaky ReLU activation. It turns out that such upsampling blocks are able to avoid checkerboard artifacts generated by transposed convolution~\cite{checkerboardArtifacts}. 

\subsection{Loss components}
    Most current approaches for depth estimation are trained using ground-truth depth maps. However, the acquisition of ground-truth depth maps in the real world is difficult. In the KITTI dataset, the ground-truth depth map is semi-dense, covering 30\% of the image, which is generated by enforcing consistency between laser scans and stereo reconstruction~\cite{sparsityCNN}. Our proposed training process is self-supervised, so no ground-truth depth map is required. The total loss consists of three components:
    \begin{equation}
    Loss = \alpha \times L_{depth} + \beta \times L_{photometric} + \gamma \times L_{smooth}
    \end{equation}

    \begin{itemize}
    \item
    Sparse Depth Loss ($L_{depth}$): The goal of depth completion is to fill in the depth on pixels where there is no valid depth. The pixels where the input sparse depth map has a valid value should remain unchanged during the process. The $L_{depth}$ penalizes the difference on pixels with known depth before and after the depth completion task. The sparse depth supervision loss is defined as follows:
    \begin{equation}
        L_{depth}(pred, sparse) = || (pred - sparse)_{d>0} ||
    \end{equation}
    where $pred$ is the predicted dense depth map and $sparse$ is the input sparse depth map. In the sparse depth map, pixels greater than zero are valid. During the experiments, mean absolute error is used. $sparse$ is replaced with ground-truth depth maps if they are given.
  
    \item  
    Photometric Loss ($L_{photometric}$): Let $I_L$ and $I_R$ be the left and right images. Given the dense left disparity map $D_L$, any pixel in the left image has a corresponding pixel in the right image. Therefore, we are able to create a warping function $F(I, D)$ to synthesize the other camera's view using bilinear sampling, which is differentiable. $I_{L}' = F(I_R, D_L)$. We then penalize the visual difference between the input image and warped image. For the left side, the photometric loss is defined as follows:
    \begin{equation}
        L_{photometric}(I_{L}', I_{L}) = \lambda_1 S(I_L,I_L^{'}) + \lambda_2 |I_L - I_{L}^{'}|
    \end{equation}
    where $S()$ is the structure similarity index~\cite{SSIM}, and we set $\lambda_1 = 0.85$ and $\lambda_2 = 0.2$ during experiments.

    \item  
    Smoothness Loss ($L_{smooth}$): Minimizing photometric loss tends to introduce high-frequency noise, which makes the depth map non-smooth and incorrect. The smoothness loss helps alleviate this problem. 
    This smoothness loss computes the sum of the weighted second derivative of the disparity map and depth map, and the weight is the exponential of the second derivative of the input image. For the left side, smoothness loss on depth maps is defined as follows:
    \begin{equation}
    \begin{split}
        L_{smooth}(I_L, D_L) = \frac{1}{N}\sum (|\nabla_{x}^{2}D_L|e^{-|\nabla_{x}^{2}I_L|} +\\  |\nabla_{y}^{2}D_L|e^{-|\nabla_{y}^{2}I_L|})
    \end{split}
    \end{equation}
    where $N$ is number of pixels, $\nabla_{x}^{2}$ and $\nabla_{y}^{2}$ are second derivatives along the X and Y axes. We compute smoothness loss on both disparity maps and depth maps. Since disparity is inversely proportional to depth, those two smoothness loss terms are giving importance to both close and distant objects.

    \end{itemize}

\section{EXPERIMENTAL EVALUATION}
In this section, we explain our implementation details and present qualitative and quantitative results.

\subsection{Dataset}
    The depth completion task on the KITTI dataset~\cite{sparsityCNN} provides stereo images, sparse depth maps and semi-dense ground-truth depth maps in training and validation sets, but only monocular images in the test set. The sparse depth map is generated by projecting 3D LIDAR point clouds (collected from Velodyne HDL-64E) onto the image plane. Ground-truth depth maps are generated by accumulating 11 frames of point clouds from the LIDAR. This accumulating process only produces semi-dense depth maps, and valid depth pixels only exist for the bottom part of the images. The dataset consists of 42,949 pairs of training images and 3,426 pairs of validation images with both left and right depth maps, including 1,000 cropped images with a fixed size. For convenience, we stick to take left sparse depth maps as inputs, so only half of the dataset is used.
    
    We are not able to submit results on the KITTI benchmark, since the test set does not contain stereo images. Still, we report rough evaluation results on the KITTI dataset using part of the validation set. We randomly split the validation set into two sub-sets, 2,426 pairs of stereo images for testing and 1,000 for validation.

\subsection{Implementation Details}
    The model is implemented in TensorFlow~\cite{tensorflow}. Input stereo images are normalized to values ranging from -1 to 1. Depth inversion introduced by~\cite{ku2018defense} is conducted on input sparse depth maps since it creates a buffer gap between valid and empty pixel values. All parameters are randomly initialized using truncated zero-mean Gaussian distribution. During the training, images are randomly cropped down to 256x1024 patches on the bottom of images before being fed into the network. We use the Adam optimizer~\cite{adam} with $\beta_1=0.9$, $\beta_2=0.999$ and $\epsilon=1e{-8}$. We have a batch size of 10. The learning rate starts at 1e-4 and reduces to 5e-5 after training for 6 epochs. A total of 12 epochs are run. All experiments are run on two NVIDIA Titan-X GPUs. No data augmentation is performed in the experiments.

\subsection{Evaluation}
    \begin{table*}[t!]
    \begin{center}
    \begin{tabular}{|l|c|c|c|c|}
    \hline
    Method & iRMSE(1/km) & iMAE(1/km) & RMSE(mm) & MAE(mm) \\
    \hline\hline
    SparseCNN~\cite{sparsityCNN} & 4.94 & 1.78 & 1601.33 &	481.27\\
    ADNN~\cite{chodosh18} & 59.39 &	3.19 & 1325.37 & 439.48\\
    Spade~\cite{jaritz2018sparse} & 2.60 & 0.98 & 1035.29 & 248.32\\
    CSPN~\cite{cheng2018depth} & 2.93 & 1.15 & 1019.64 & 279.46 \\
    Sparse2dense~\cite{maSelf} & 2.80 &	1.21 &	814.73 & 249.95\\
    RGB\_guide\&Certainty~\cite{van2019sparse}& 3.26 & \textbf{0.74} & \textbf{778.89} & \textbf{168.82}\\
    \hline
    LiMono(test-set + with GT) & 3.49 & 1.73 & 1058.06 & 371.86\\
    \hline
    LiMono(split-test-set + with GT) & 3.88 & 2.53 & 1178.22 & 372.04\\
    LiStereo(split-test-set + with GT) & \textbf{2.19} & 1.10 & 832.16 & 283.91\\
    \hline
    \end{tabular}
    \end{center}
    \caption{Comparison with other supervised methods on KITTI benchmark depth completion task. `LiStereo' refers to proposed model taking stereo images as inputs. `LiMono' refers to the model taking monocular images. `test-set' refers to results that are reported on KITTI test set. `split-test-set' refers to results that are reported on split test set from KITTI validation set introduced in Section 4.1. `with GT' refers to training model using ground-truth label. With stereo information, our proposed method outperforms models with only monocular information.     }
    \label{results_kitti}
    \end{table*}

    \begin{table*}[t!]
    \begin{center}
    \begin{tabular}{|c|c|c|c|c|}
    \hline
    Method & iRMSE(1/km) & iMAE(1/km) & RMSE(mm) & MAE(mm) \\
    \hline\hline
    Sparse2dense(split-test-set + w/o GT)~\cite{maSelf} & 4.08 & 1.61 & 1301.05 & 352.22\\
    Sparse2dense(test-set + w/o GT)~\cite{maSelf} & 4.07 & 1.57 & 1299.85 & 350.32\\
    \hline
    LiStereo(split-test-set + w/o GT) & \textbf{3.83} & \textbf{1.32} & \textbf{1278.87} & \textbf{326.10}\\
    \hline
    \end{tabular}
    \end{center}
    \caption{Comparison with other self-supervised methods on KITTI benchmark depth completion task. `LiStereo' refers to proposed model taking stereo images as inputs. `test-set' refers to results that are reported on KITTI test set. `split-test-set' refers to results that are reported on split test set from KITTI validation set introduced in Section 4.1. `w/o GT' refers to training model in self-supervised training manner. With stereo information, our proposed method outperforms Sparse2Dense which is trained using temporal frames. Similar results of Sparse2Dense on split-test-set and test-set justify our strategy of split validation set introduced in Section 4.1.  
    }
    \label{results_kitti_self}
    \vspace{-3mm}
    \end{table*}

    We evaluate the proposed method on KITTI dataset. The main metric we use in this paper is root-mean-square error (RMSE in mm), which computes the L2 norm difference on all pixels where the ground-truth depth map is valid. We also report results of other metrics, such as mean-average error (MAE in mm), inverse mean-average error (iMAE in 1/km) and inverse-root-mean-square error (iRMSE in 1/km). 

    The evaluation results are listed in Table \ref{results_kitti} and Table \ref{results_kitti_self} for supervised and self-supervised approaches respectively. Because no stereo images are provided in test set of KITTI dataset, we devise a strategy to roughly compare with approaches of the submitted results on the KITTI benchmark depth completion task leaderboard. In contrast to LiStereo, we have built a new model and call it LiMono, which only takes monocular images as inputs. It serves as the baseline. The LiMono is created by removing the right branch of the Siamese structure as shown in the Figure \ref{architecture}, and it is trained using ground-truth label. We submitted results of LiMono to the KITTI benchmark. The LiMono is able to achieve state-of-the-art performance. We also report results of both the LiMono and LiStereo in the table using the test set described in Section 4.1. With stereo information, LiStereo trained with ground-truth labels has a much lower RMSE than the LiMono does. This demonstrates that stereo cues have advantage over monocular cues and are able to improve the accuracy.

\subsection{Ablation Study on Weights of Loss}
    The photometric loss is key for the model to learn to fill depth values in empty pixels and to be trained in a self-supervised manner. Here, we study the influence of the weight of the photometric loss on final results. We experiment with setting the photometric loss weight $\beta$ between 0 and 2, and the results are shown in Table \ref{study_loss_weights}. Numbers are reported on the validation set described in Section 4.1. This table demonstrates that photometric loss does help complete depth on sparse input, reducing RSME from 1970mm to 1277mm with $\beta = 0$ and $\beta = 0.5$ respectively. However, too high a weight on photometric loss decreases performance. This is most likely due to the high weight on regions in which the photometric loss is poorly defined like those with low texture or specular reflections. So we set $\beta = 0.5$ throughout experiments. Besides, we set $\gamma$ = 0.001 for models which are trained using ground-truth labels, since smoothness loss will over-smooth edges between objects.

    \begin{table*}[t]
    \begin{center}
    \footnotesize
    \begin{tabular}{|l|c|c|c|c|c|c|c|c|}
    \hline
    \specialcell{Loss \\ Weights} & \specialcell{$\alpha$ = 1\\ $\gamma$ = 0.01\\ $\beta$ = 0.0} & \specialcell{$\alpha$ = 1\\ $\gamma$ = 0.01\\ $\beta$ = 0.2} & \specialcell{$\alpha$ = 1\\ $\gamma$ = 0.01\\ $\beta$ = 0.5} & \specialcell{$\alpha$ = 1\\ $\gamma$ = 0.01\\ $\beta$ = 0.8} & \specialcell{$\alpha$ = 1\\ $\gamma$ = 0.01\\ $\beta$ = 1} & \specialcell{$\alpha$ = 1.0\\ $\gamma$ = 0.01\\ $\beta$ = 1.5} & \specialcell{$\alpha$ = 1\\ $\gamma$ = 0.01\\ $\beta$ = 2.0} & \specialcell{$\alpha$ = 1\\ $\gamma$ = 0.001\\ $\beta$ = 0.5} \\
    \hline
    RMSE(mm) & 1970.63 & 1522.03 & \textbf{1277.36} & 1388.58 & 1356.96 & 1344.96 & 1434.89 & 1424.01 \\
    \hline
    \end{tabular}
    \end{center}
    \caption{Ablation Study on Weights of Loss. Photometric loss does help complete depth on sparse input, reducing RMSE from 1970mm to 1277mm with $\beta = 0$ and $\beta = 0.5$ respectively. However, too much weight on photometric loss makes the results worse. The model with $\alpha$ = 1, $\gamma$ = 0.01 and $\beta$ = 0.5 achieves the lowest RMSE.}
    \label{study_loss_weights}
    \vspace{-3mm}
    \end{table*}

\subsection{Ablation Study on Sparsity-invariant Convolutions}
     The sparsity-invariant operations are designed for extracting features from sparse input data, so they are suitable for the depth completion task. In this study we replace all convolution layers within LIDAR branch with the sparsity-invariant convolution introduced by~\cite{sparsityCNN} and also remove all batch normalization layers. Quantitative results are shown in Table \ref{sparseConv}. Numbers are reported on the test set described in Section 4.1. In the table, `conv' refers to model using regular convolution layer. `sparse conv' refers to model in which regular convolution layers are replaced with sparsity-invariant convolution layers in LIDAR branch. `with GT' refers to model trained using ground-truth labels. `w/o GT' refers to model trained in a self-supervised manner. It shows slightly better results of regular convolution than sparsity-invariant convolution in terms of RMSE. So the main results of the paper employ regular convolution. 
 
    \begin{table*}[t!]
    \begin{center}
    \footnotesize
    \begin{tabular}{|l|c|c|c|c|}
    \hline
    Method  & iRMSE(1/km) & iMAE(1/km) & RMSE(mm) & MAE(mm) \\
    \hline\hline
    LiStereo (sparse conv) + w/o GT & 4.12 & 1.56 & 1379.31 & 355.42\\
    LiStereo (conv) + w/o GT & 3.83 & 1.32 & 1278.87 & 326.10\\
    LiStereo (sparse conv) + with GT & 2.22 & \textbf{1.02} & 894.56 & \textbf{268.69}\\
    LiStereo (conv) + with GT & \textbf{2.19} & 1.10 & \textbf{832.16} & 283.91\\
    \hline
    \end{tabular}
    \end{center}
    \caption{Ablation study on regular convolution and sparsity-invariant convolution. All models take stereo images as inputs. `conv' refers to model using regular convolution layer. `sparse conv' refers to model in which regular convolution layers are replaced with sparsity-invariant convolution layers in LIDAR branch. `with GT' refers to model trained using ground-truth label. `w/o GT' refers to model trained in a self-supervised manner. Results are reported on split test set introduced in section 4.1. It shows slightly better results of regular convolution than sparsity-invariant convolution in terms of RMSE.}
    \label{sparseConv}
    \vspace{-5mm}
    \end{table*}

\subsection{Analysis on sparsity of inputs during training}
\begin{figure*}[t]
        \centering
        \includegraphics[width=1\linewidth]{{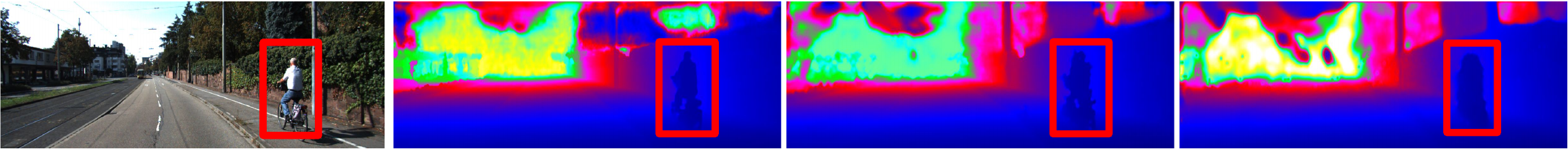}}
        \caption{Qualitative results on different levels of input sparsity. The model is trained in a self-supervised manner. From left to right: corresponding left color image, predicted dense depth map from original sparse depth map, predicted dense depth map from input of sparsity level 0.1 and predicted dense depth map from input of sparsity level of 0.01. More details are shown with denser inputs.}
        \label{fig:visual_results}
        \vspace{-4mm}
    \end{figure*}

\begin{figure*}[t!]
    \begin{center}
    \begin{tabular}{cc}
    \hbox{\includegraphics[width=0.4\linewidth,height=0.22\linewidth]{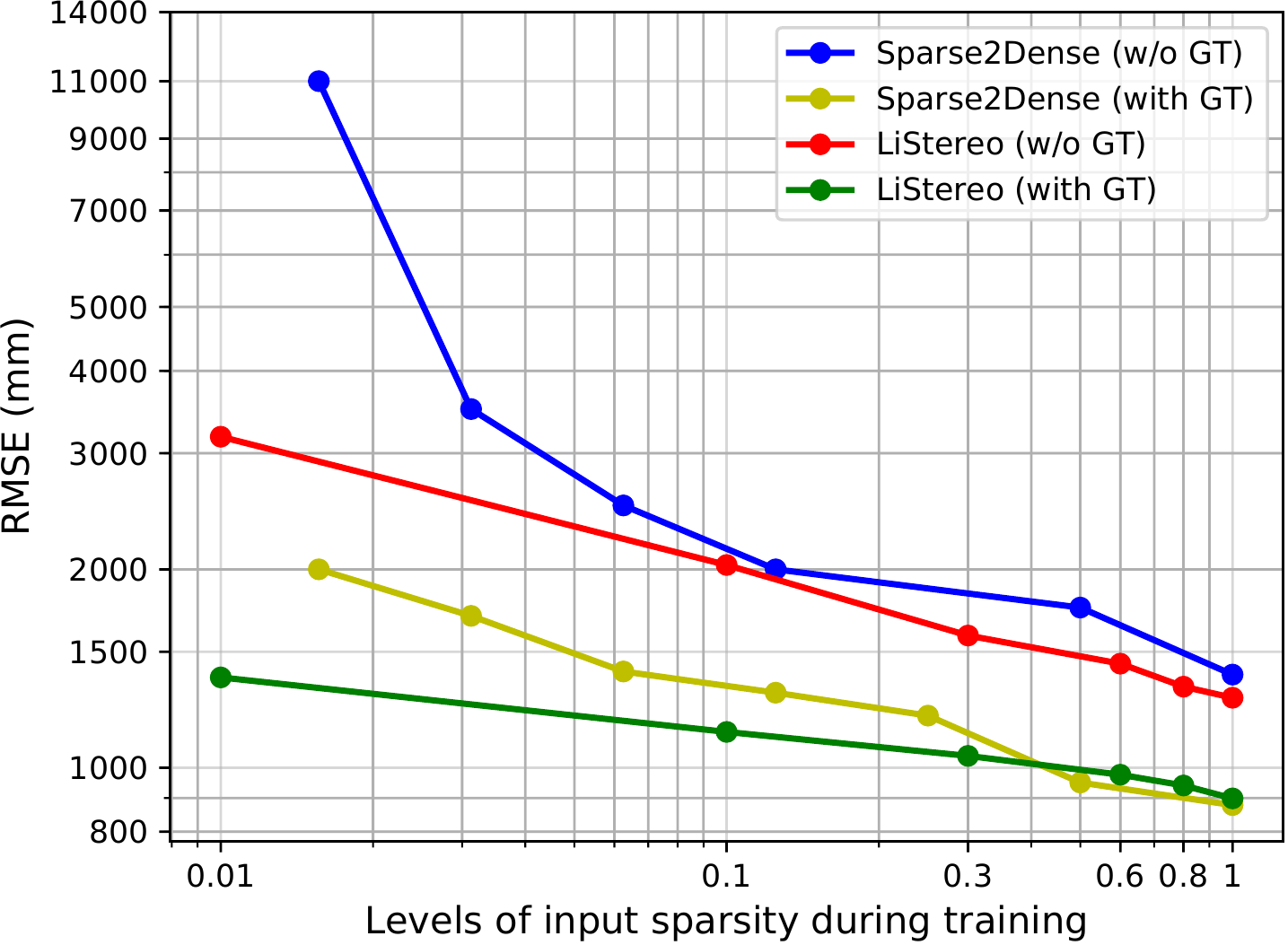}}&
    \hbox{\includegraphics[width=0.4\linewidth,height=0.22\linewidth]{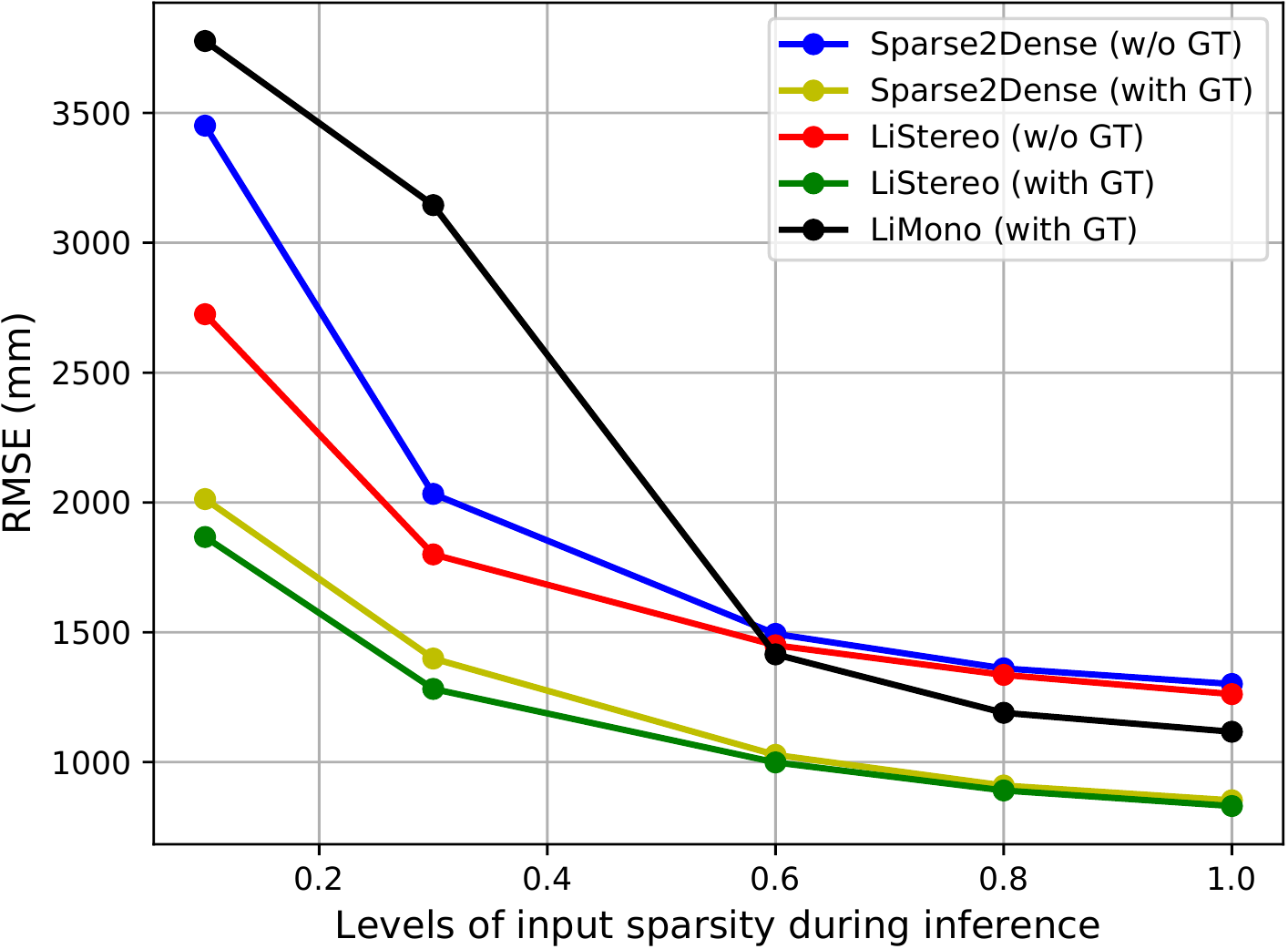}}\\
    (a)&(b)
    \end{tabular}
    \caption{Errors on different levels of input sparsity during training and inference. The X axis represents level of input sparity, and the Y axis represents RMSE (mm). `LiStereo' refers to our proposed model. `LiMono' refers to the model taking monocular images as inputs. `Sparse2Dense' refers to model proposed by \cite{maSelf}. `w/o GT' means that the model is trained in a self-supervised manner and `with GT' means that the model is trained using ground-truth label. (a) Errors during the training. Results are evaluated on split validation set introduced in Section 4.1. For each data point in the figure, the model is trained and evaluated using certain input sparsity specified by X axis; (b) Errors during the inference. Models are evaluated on the split validation set introduced in Section 4.1. We train the model using original depth maps but feed it with different sparsity levels of input depth maps during inference. Results of Sparse2dense is acquired by using the published pretrained model by \cite{maSelf}. Both figures show that our proposed model is robust to sparse input and has the potential to take depth maps generated from low-resolution LIDARs.}
    \label{fig:input_sparsity}
    \end{center}
    \vspace{-8mm}
    \end{figure*}

    In this section, we investigate the influence of different sparsity of input depth maps by uniformly subsampling valid pixels in the input depth maps. This simulates situations where low-resolution LIDARs are used. We do analysis on both supervised and self-supervised versions of the proposed model. Qualitative results of self-supervised model are shown in Figure \ref{fig:visual_results}. During this study, we choose the self-supervised model ($\alpha$ = 1, $\beta$ = 0.5 and $\gamma$ = 0.01) and the supervised model ($\alpha$ = 1, $\beta$ = 0.5 and $\gamma$ = 0.001). Models are trained and evaluated using different levels of sparsity of input depth maps and we report quantitative results in the Table \ref{study_sparity}. Level of sparsity (LoS) in the table means the ratio of valid pixels sampled from the original depth map, and RMSEs are reported. As expected Level of sparsity (LoS) in the table means the ratio of valid pixels sampled from the original depth map. We also compare our model with both the supervised and self-supervised versions of Sparse2Dense~\cite{maSelf}, which were trained using monocular images and temporal frames. Results are shown in Figure \ref{fig:input_sparsity} (a). Stereo cues turn out to be more robust to sparse input than monocular information in both supervised and self-supervised training process. Note the flatter error curve of for our model compared with that of Sparse2Dense. Especially, with only 1 percent density of the original sparse map, the RMSE of our self-supervised model is 3177.83mm, which is much smaller than around 11000mm achieved by self-supervised version of Sparse2Dense. 

    \vspace{-2mm}
    \begin{table*}[t]
    \begin{center}
    \footnotesize
    \begin{tabular}{|c|c|c|c|c|c|c|}
    \hline
    Level of sparsity & 0.01 & 0.1 & 0.3 & 0.6 & 0.8 & 1 \\
    \hline
    LiStereo (self-supervised) & 3177.83 & 2030.36 & 1587.28 & 1438.79 & 1327.68 & \textbf{1277.36} \\
    \hline
    LiStereo (supervised) & 1371.28 & 1133.58 & 1042.62 & 976.35 & 940.25 & \textbf{898.77} \\
    \hline
    \end{tabular}
    \end{center}
    \caption{Errors on different levels of input sparsity during training. RMSE (mm) from the supervised and the self-supervised models are reported. Level of sparsity (LoS) in the table means the ratio of valid pixels sampled from the original depth map. }
    \label{study_sparity}
    \end{table*}

\subsection{Analysis on sparsity of inputs during inference}
\vspace{-2mm}
    We conduct an analysis on different sparsity levels of input depth maps during inference. In this case, the model is trained using original sparse depth maps but is provided different input sparsity levels during inference. This simulates situations where a primary model, which is trained and fed with dense data, is deployed to other applications in which only the low-resolution LIDAR is provided. The results is shown in the Figure \ref{fig:input_sparsity} (b). Results are reported on the validation set described in Section 4.1. We also report results of Sparse2Dense and LiMono. The LiMono model turns out to have the highest and steepest error curve, which means that training with temporal information used in Sparse2Dense or stereo cues used in our proposed model enable the model to be better to exploit information in the color image. Compared with Sparse2Dense, LiStereo trained in either supervised or self-supervised manner is more robust in terms of input sparsity, which means stereo images are able to provide more information than monocular images, even if multiple frames of monocular images are used during the training.

\section{CONCLUSIONS}
In this paper, we propose a model to upsample sparse depth map generated from LIDAR using stereo cues. Depth maps from LIDAR are accurate but sparse and stereo estimation generates less reliable depth maps which are dense. Our proposed method combines these two approaches together making use of the advantages of both to generate accurate high-resolution depth maps, which is important for autonomous driving and other robotic applications. Compared to using monocular images as the guidance, stereo cues turn out to be more robust to highly sparse inputs and so they reduces LIDAR resolution requirement. Thus, we can use the LIDAR with lower resolution and still produce comparable results as a high-resolution LIDAR. The high-resolution LIDAR (Velodyne 128) is more than ten times of the cost of the low-resolution LIDAR (Velodyne 16) and stereo systems. So the proposed method would offer massive cost reduction for high resolution depth estimation. 

\section*{ACKNOWLEDGMENT}
This work was supported by a grant from Ford Motor Company via the Ford-UM Alliance under award N022884.

\printbibliography

\end{document}